\documentclass{article}

\usepackage{microtype}
\usepackage{graphicx}
\usepackage{subfigure}
\usepackage{booktabs} 
\usepackage{todonotes}
\usepackage{arydshln}

\usepackage{hyperref}

\usepackage[accepted]{icml2020}

\icmltitlerunning{Non-Pharmaceutical Intervention Discovery}

\begin{document}

\twocolumn[
\icmltitle{Non-Pharmaceutical Intervention Discovery with Topic Modeling
}




\begin{icmlauthorlist}

\icmlauthor{Jonathan Smith}{l6}
\icmlauthor{Borna Ghotbi}{vi,ubc}
\icmlauthor{Seungeun Yi}{l6}
\icmlauthor{Mahboobeh Parsapoor}{vi,ut}

\end{icmlauthorlist}

\icmlaffiliation{l6}{Layer 6 AI, Canada}
\icmlaffiliation{vi}{Vector Institute, Canada}
\icmlaffiliation{ut}{University of Toronto, Canada}
\icmlaffiliation{ubc}{University of British Columbia, Canada}

\icmlcorrespondingauthor{Jonathan Smith}{jonathan@layer6.ai}

\icmlkeywords{Machine Learning, ICML}

\vskip 0.3in
\setlength{\parskip}{0.5em} 
]



\printAffiliationsAndNotice{}  

\begin{abstract}
We consider the task of discovering categories of non-pharmaceutical interventions during the evolving COVID-19 pandemic. We explore topic modeling on two corpora with national and international scope. These models discover existing categories when compared with human intervention labels while reduced human effort needed.
\end{abstract}

\vspace{-24pt}

\section{Introduction}

The COVID-19 pandemic has seen varying government responses in countries around the world. More than 2.24 million confirmed cases and 152.5 thousand deaths worldwide (January 1, 2020 - April 19, 2020)~\cite{WHO} are attributed to novel coronavirus (SARS-CoV-2). Due to the high transmission rate, non-pharmaceutical interventions (NPIs) play a critical role in stemming the spread of the virus~\cite{ferguson2020report}.
Studies assess the effectiveness of interventions in "flattening the curve"~\cite{koo2020interventions, tuite2020mathematical}, using individual or compartmental models~\cite{biswas2014seir} to simulate scenarios for fixed parameters. Broad categories encompassing physical distancing, business closures, and travel restrictions, are used and tailored to the specific country or region the model focuses on~\cite{wang2020evolving}. More specific interventions are less likely to be tracked, leaving the effects of increased funding, stratified closure, and administrative flexibility unestimated. Increasing the granularity of NPIs recorded is limited by manual labor involved in intervention identification and evaluation. This is seen in the categorizations used by four projects currently compiling data~\cite{mccoy2020cannpi, hitcovid, acaps2020, hale2020oxf} which record 63, 23, 35, and 13 NPI types respectively.

In this study we leverage the availability of global text data on intervention announcements to automate the task of intervention category discovery, which will speed up the identification of interventions with strong effects on transmission.

\begin{table}[H]

\caption{\label{tab:top-words}Top 10 words per topic for 3 topics in $D_{CAN}$, using LDA with $K=25$.}
\vspace{6pt}

\begin{tabular}{llllll}
\hline
Topic & Words \\ \hline
0  & businesses, financial, program, government,\\
   &  business, relief, time, covid, announced, fees \\ \hdashline 
11 & closed, closure, facilities, parks, public,\\
   & closures, visitors, centres, please\_visit, notice \\ \hdashline
20 & park, leash, couple, equipment, play, enjoy, \\
   & neighbourhood, staying, bylaws, however \\ \hline
\end{tabular}
\end{table}

We propose a method for using topic modeling to discover intervention categorizations. This method uses two well-known topic modeling algorithms, Latent Dirichlet Allocation (LDA) and  Hierarchical Dirichlet Processes (HDP), combined with the unique structure of the NPI category discovery problem. We introduce evaluation methods for comparing unsupervised discovery models with existing intervention labels and demonstrate that our approach discovers known categories of NPIs as seen in the selected topics in Table \ref{tab:top-words}. Our method has shown useful performance on datasets with global scope, suggests that better categorizations can be learned by pooling knowledge across countries.

\section{Datasets, Method, Results}

\subsection{Datasets}

This paper uses two datasets: 1) the Oxford COVID-19 Government Response Tracker v4.0~\cite{hale2020oxf} ($D_{OXF}$), which contains 13 intervention categories from 156 countries, and 2) CAN-NPI dataset~\cite{mccoy2020cannpi} ($D_{CAN}$), which contains 63 intervention categories found in Canada. The former provides notes for each change to the status of an intervention citing public announcement texts, quotes, email interactions with public health authorities, and links. In contrast, the latter provides more standardized full-text articles for each intervention tagged at the provincial, territorial, and municipal level in a single country. The $D_{CAN}$ dataset also links all eligible interventions from the Oxford intervention categories. For consistency of results, the period of January 1, 2020 - April 19, 2020, is used for both of the datasets. Table~\ref{tab:datasets} compares dataset size and composition. It is also essential that international data is evenly sampled from across regions. In $D_{OXF}$, countries have similar distributions of average note counts in Asia (31), Europe (28), Africa (29), and the Americas (29).

\begin{table}[H]

\caption{\label{tab:datasets} Summary of datasets: The number of documents $D$, vocabulary size $V$, average words per document $W_{mean}$, and standard deviation in words per document $W_{std}$.}
\vspace{6pt}

\begin{tabular}{lllll}
\hline
Dataset & $D$ & $V$ & $W_{mean}$ & $W_{std}$ \\ \hline
$D_{CAN}$ & 1,290 & 574,417     & 248       & 204    \\ 
$D_{OXF}$ & 3,900 & 114,743     & 17        & 16     \\  \hline
\end{tabular}
\end{table}

\subsection{Method}
\subsubsection{Preprocessing and Topic Modeling}
A corpus is constructed from each dataset. We preprocess the documents in each dataset to remove default english stopwords from NTLK~\cite{ntlk2009} and geographic words (eg. province or country names) from data labels. Bigrams are constructed from word pairs that generally occur together (eg. “public health”) and are added as well.

We apply LDA~\cite{blei2012probabilistic} and HDP~\cite{teh2004hierarchical} using \texttt{gensim}~\cite{rehurek_lrec} to generate topics for each corpus. Topic coherence, $C_v$, is calculated as a measure of topic interpretability using co-occurence of words and cosine similarity between each top word and the total of all top words~\cite{roder2015exploring,syed2017full}. LDA is run to generate a specific number of topics, $K$, then coherence is calculated for each $K$. For LDA, the maximum coherence is associated with the near-optimal number of topics found from a corpus. HDP predetermines the number of topics and calculates a single coherence over all of them. The online nature of these algorithms allows for topics that can be easily extended when more data is gathered, helping researchers respond quickly in a fast-moving pandemic.

\subsubsection{Similarity Evaluation}
We evaluate the suitability of a topic as an intervention category by comparing with existing labels from subject matter experts (SMEs). Mean cosine similarity ($S$) between weighted topic occurrence per document ($\textbf{\textit{T}} \in [0,1]^{D \times T}$) and label occurrence ($\textbf{\textit{L}} \in [0,1]^{D \times C}$) for $D$ documents, $C$ labels, and $T$ topics is one metric. This is calculated as:

$$ \frac{1}{C T} \sum_{i=1}^{T} \sum_{j=1}^{C} (\textbf{\textit{T}}^{\top} \textbf{\textit{L}})_{i,j} $$

A higher similarity shows that topics are similar in structure to given labels. We evaluate individual label discovery by assuming that each topic can correspond to at least one intervention category. To determine the coverage ($Cov$) we select the topic with the highest similarity for each label and count the number of distinct topics used to cover all labels. Coverage ratio ($Cov(\%)$) is found with $Cov / C$. Coverage ratio combined with high similarity indicates the discovery of label categories. Combining these metrics enables model performance evaluation without additional work by SMEs.

\subsection{Results}

\vspace{-6pt}

\begin{table}[]

\caption{\label{tab:topic-perf}Evaluation of topic coherence ($C_v$), mean cosine similarity between topics and intervention categories ($S$), and coverage ratio of top 1 topic selection over intervention categories $Cov (\%)$ for the top $K$ topics. Two topic models (TM) are compared.}
\vspace{6pt}
\begin{tabular}{llllll}
\hline
Dataset & TM &  $K$ & $C_v$ & $S$ & $Cov (\%)$  \\ \hline
$D_{CAN}$ & LDA & 10    & 0.435 & \textbf{0.029} & 0.17    \\
          & LDA & 25    & 0.451 & 0.023         & 0.33   \\
          & LDA & 50    & 0.427 & 0.023         & 0.44   \\
          & LDA & 100   & 0.379 & 0.018         & 0.63   \\\hdashline
          & HDP & 10    & \textbf{0.608} & \textbf{0.029}   & 0.17     \\
          & HDP & 25    & \textbf{0.608} & 0.025      & 0.37  \\
          & HDP & 50    & \textbf{0.608} & 0.019      & 0.59  \\
          & HDP & 100   & \textbf{0.608} & 0.013         & \textbf{0.68}   \\ \hline
$D_{OXF}$ & LDA & 10    & 0.505 & 0.010 & 0.64    \\
          & LDA & 25    & 0.427 & 0.010 & 0.86   \\
          & LDA & 50    & 0.436 & 0.010  & 0.93   \\
          & LDA & 100   & 0.373 & 0.008   & 0.93  \\\hdashline
          & HDP & 10    & \textbf{0.683} & \textbf{0.141}      & 0.64     \\
          & HDP & 25    & \textbf{0.683} & 0.137      & 0.93  \\ 
          & HDP & 50    & \textbf{0.683} & 0.128      & 0.93  \\
          & HDP & 100   & \textbf{0.683} & 0.114 & \textbf{1.00}   \\ \hline

\end{tabular}

\end{table}

We summarize the results of category discovery with topic modeling on $D_{CAN}$ and $D_{OXF}$ datasets in Table \ref{tab:topic-perf} using $K \in \{10, 25, 50, 100\}$. HDP leads to topics with significantly more coherence, while a clear tradeoff must be made between overall similarity and coverage. Similarity drops less precipitously as the number of topics increases when using LDA as opposed to HDP. $D_{OXF}$ has a much smaller number of categories, so there is a clear early convergence in coverage. In practice, a higher number of topics will be harder for humans to check for interventions but should lead to coverage of existing categories and aid with discovery.

\section{Conclusion and Future Work}
In this paper, we briefly demonstrate LDA- and HDP-based methods for NPI discovery on $D_{CAN}$ and $D_{OXF}$. As our results have verified, combining coherence, topic similarity, and coverage can provide provide intervention category suggestions for experts to explore and find new NPIs leveraging national and international data. For future work, we will explore prediction-focused supervised LDA and identification of NPIs in an online manner.


\bibliography{example_paper}

\begin{thebibliography}{16}
\providecommand{\natexlab}[1]{#1}
\providecommand{\url}[1]{\texttt{#1}}
\expandafter\ifx\csname urlstyle\endcsname\relax
  \providecommand{\doi}[1]{doi: #1}\else
  \providecommand{\doi}{doi: \begingroup \urlstyle{rm}\Url}\fi

\bibitem[WHO(2020)]{WHO}
{COVID-19 situation report-19}.
\newblock
  \url{http://aiweb.techfak.uni-bielefeld.de/content/bworld-robot-control-software/},
  2020.
\newblock [Online; accessed 12-May-2020].

\bibitem[ACAPS(2020)]{acaps2020}
ACAPS.
\newblock {ACAPS} {COVID-19}: Government measures dataset - humanitarian data
  exchange (version 1.0).
\newblock 2020.
\newblock
  \url{https://data.humdata.org/dataset/acaps-covid19-government-measures-dataset}.

\bibitem[Bird et~al.(2009)Bird, Klein, and Loper]{ntlk2009}
Bird, S., Klein, E., and Loper, E.
\newblock \emph{Natural Language Processing with Python}.
\newblock O’Reilly Media Inc, 2009.
\newblock URL \url{http://nltk.org/book}.

\bibitem[Biswas et~al.(2014)Biswas, Paiva, and De~Pinho]{biswas2014seir}
Biswas, M.~H., Paiva, L.~T., and De~Pinho, M. d.~R.
\newblock A seir model for control of infectious diseases with constraints.
\newblock \emph{MBE}, 2014.
\newblock \doi{10.3934/mbe.2014.11.761}.

\bibitem[Blei(2012)]{blei2012probabilistic}
Blei, D.~M.
\newblock Probabilistic topic models.
\newblock \emph{Communications of the ACM}, 55\penalty0 (4):\penalty0 77--84,
  2012.

\bibitem[Ferguson et~al.(2020)Ferguson, Laydon, Nedjati~Gilani, Imai, Ainslie,
  Baguelin, Bhatia, Boonyasiri, Cucunuba~Perez, Cuomo-Dannenburg,
  et~al.]{ferguson2020report}
Ferguson, N., Laydon, D., Nedjati~Gilani, G., Imai, N., Ainslie, K., Baguelin,
  M., Bhatia, S., Boonyasiri, A., Cucunuba~Perez, Z., Cuomo-Dannenburg, G.,
  et~al.
\newblock Report 9: Impact of non-pharmaceutical interventions (npis) to reduce
  covid19 mortality and healthcare demand.
\newblock 2020.

\bibitem[Hale et~al.(2020)Hale, Petherick, Phillips, and Webster]{hale2020oxf}
Hale, T., Petherick, A., Phillips, T., and Webster, S.
\newblock Variation in government responses to covid-19 (version 4.0).
\newblock 2020.
\newblock \doi{BSG-WP-2020/031}.

\bibitem[HIT~COVID(2020)]{hitcovid}
HIT~COVID, T.
\newblock Health interventions tracking for covid-19 (hit-covid) (version
  v0.3).
\newblock 2020.
\newblock \doi{10.5281/zenodo.3765628}.

\bibitem[Koo et~al.(2020)Koo, Cook, Park, Sun, Sun, and
  Lim]{koo2020interventions}
Koo, J.~R., Cook, A.~R., Park, M., Sun, Y., Sun, H., and Lim, J. T. e.~a.
\newblock Interventions to mitigate early spread of sars-cov-2 in singapore: a
  modelling study.
\newblock \emph{The Lancet Infectious Diseases}, 2020.
\newblock \doi{https://doi.org/10.1016/S1473-3099(20)30162-6}.

\bibitem[McCoy et~al.(2020)McCoy, Smith, Anchuri, Berry, Pineda, Harish, Lam,
  Yi, Hu, Canadian Open Data Working~Group, and Fine]{mccoy2020cannpi}
McCoy, L.~G., Smith, J., Anchuri, K., Berry, I., Pineda, J., Harish, V., Lam,
  A.~T., Yi, S.~E., Hu, S., Canadian Open Data Working~Group, N.-P.~I., and
  Fine, B.
\newblock {CAN}-{NPI}: A curated open dataset of canadian non-pharmaceutical
  interventions in response to the global covid-19 pandemic.
\newblock Preprint at
  \url{https://www.medrxiv.org/content/early/2020/04/22/2020.04.17.20068460},
  2020.

\bibitem[{\v R}eh{\r u}{\v r}ek \& Sojka(2010){\v R}eh{\r u}{\v r}ek and
  Sojka]{rehurek_lrec}
{\v R}eh{\r u}{\v r}ek, R. and Sojka, P.
\newblock {Software Framework for Topic Modelling with Large Corpora}.
\newblock In \emph{{Proceedings of the LREC 2010 Workshop on New Challenges for
  NLP Frameworks}}, pp.\  45--50, Valletta, Malta, May 2010. ELRA.
\newblock \url{http://is.muni.cz/publication/884893/en}.

\bibitem[R{\"o}der et~al.(2015)R{\"o}der, Both, and
  Hinneburg]{roder2015exploring}
R{\"o}der, M., Both, A., and Hinneburg, A.
\newblock Exploring the space of topic coherence measures.
\newblock In \emph{Proceedings of the eighth ACM international conference on
  Web search and data mining}, pp.\  399--408, 2015.

\bibitem[Syed \& Spruit(2017)Syed and Spruit]{syed2017full}
Syed, S. and Spruit, M.
\newblock Full-text or abstract? examining topic coherence scores using latent
  dirichlet allocation.
\newblock In \emph{2017 IEEE International conference on data science and
  advanced analytics (DSAA)}, pp.\  165--174. IEEE, 2017.

\bibitem[Teh et~al.(2004)Teh, Jordan, Beal, and Blei]{teh2004hierarchical}
Teh, Y.~W., Jordan, M.~I., Beal, M.~J., and Blei, D.~M.
\newblock Hierarchical dirichlet processes.
\newblock 2004.

\bibitem[Tuite et~al.(2020)Tuite, Fisman, and Greer]{tuite2020mathematical}
Tuite, A.~R., Fisman, D.~N., and Greer, A.~L.
\newblock Mathematical modelling of covid-19 transmission and mitigation
  strategies in the population of ontario, canada.
\newblock \emph{CMAJ}, 2020.
\newblock \doi{10.1503/cmaj.200476}.

\bibitem[Wang et~al.(2020)Wang, Li, Hao, Guo, Wang, Huang, He, Yu, Lin, Pan,
  Wei, and Wu]{wang2020evolving}
Wang, C.~W., Li, L., Hao, X., Guo, H., Wang, Q., Huang, J., He, N., Yu, H.,
  Lin, X., Pan, A., Wei, S., and Wu, T.
\newblock Evolving epidemiology and impact of non-pharmaceutical interventions
  on the outbreak of coronavirus disease 2019 in wuhan, china.
\newblock \emph{JAMA}, 2020.
\newblock \doi{10.1001/jama.2020.6130}.

\end{thebibliography}
\bibliographystyle{icml2020}

\end{document}